\documentclass{bmvc2k}
\usepackage{comment}
\usepackage{graphics}
\usepackage{amssymb}
\usepackage[ruled]{algorithm2e}
\usepackage{setspace}
\usepackage{multirow}
\usepackage{booktabs}
\usepackage{pifont}
\usepackage{float}
\usepackage[rightcaption]{sidecap}
\usepackage{multicol}
\floatstyle{ruled}

\newcommand\blfootnote[1]{%
	\begingroup
	\renewcommand\thefootnote{}\footnote{#1}%
	\addtocounter{footnote}{-1}%
	\endgroup
}

\title{Domain Adaptation Regularization for Spectral Pruning}

\addauthor{Laurent Dillard}{laurent.dillard@gmail.com}{1,$\dagger$}
\addauthor{Yosuke Shinya}{yosuke.shinya.j7r@jp.denso.com}{2}
\addauthor{Taiji Suzuki}{taiji@mist.i.u-tokyo.ac.jp}{3,4}

\addinstitution{
	Behold.ai\\
	New York, USA
}
\addinstitution{
	DENSO CORPORATION\\
	Tokyo, Japan
}
\addinstitution{
	The University of Tokyo\\
	Tokyo, Japan
}
\addinstitution{
	AIP-RIKEN\\
	Tokyo, Japan
}

\newcommand{\argmaxunder}[1]{\underset{#1}{\operatorname{arg\,max}}}

\newcommand{\argmax}{\mathop{\mathrm{arg\,max}}}

\def\OurAlgo{MoMaSP\xspace}
\def\OurAlgoFull{Moment Matching Spectral Pruning\xspace}

\newcommand{\E}{\mathrm{E}}

\newcommand{\mell}{m_l}

\newcommand{\Real}{\mathbb{R}}
\newcommand{\Eqref}[1]{Eq. \eqref{#1}}
\newcommand{\Tr}{\mathrm{Tr}}

\newcommand{\cmark}{\ding{51}}%

\def\AppendixSection{Supplementary Material\xspace}

\runninghead{Dillard, Shinya, Suzuki}{\OurAlgoFull}

\begin{document}

\maketitle
\blfootnote{
	\hspace{-17pt} $\dagger$Work done as an intern at AIP-RIKEN.}
\vspace{-15pt}

\setlength{\abovedisplayskip}{2pt}
\setlength{\belowdisplayskip}{2pt}
\setlength{\textfloatsep}{5pt}

\begin{abstract}
   Deep Neural Networks (DNNs) have recently been achieving state-of-the-art performance on a variety of computer vision related tasks.
   However, their computational cost limits their ability to be implemented in embedded systems with restricted resources or strict latency constraints.
   Model compression has therefore been an active field of research to overcome this issue.
   Additionally, DNNs typically require massive amounts of labeled data to be trained.
   This represents a second limitation to their deployment.
   Domain Adaptation (DA) addresses this issue by allowing knowledge learned on one labeled source distribution to be transferred to a target distribution, possibly unlabeled.
   In this paper, we investigate on possible improvements of compression methods in DA setting.
   We focus on a compression method that was previously developed in the context of a single data distribution and show that, with a careful choice of data to use during compression and additional regularization terms directly related to DA objectives, it is possible to improve compression results.
   We also show that our method outperforms an existing compression method studied in the DA setting by a large margin for high compression rates.
   Although our work is based on one specific compression method, we also outline some general guidelines for improving compression in DA setting.
\end{abstract}

\section{Introduction}

In the past years, Deep Neural Networks have attracted much attention in vision tasks.
Indeed, they led to significant improvements in many problems including image classification and object detection~\cite{AlexNet_NIPS2012_4824, Faster_R-CNN_NIPS}. This was mainly achieved by the development of complex new architectures based on Convolutional Neural Networks (CNNs)~\cite{Inception_Szegedy_2015_CVPR, VGGNet, ResNet_CVPR2016}. These architectures typically involve a huge number of parameters and operations for inference, making them ill-suited to being deployed in constrained environments. Several different strategies have been developed in order to circumvent this issue by compressing the models into smaller ones achieving similar performances. The main methods can be categorized into the following schemes: selectively pruning parameters~\cite{NIPS1988_156, ijcai2018-330}, distillation~\cite{conf/kdd/BucilaCN06, NIPS2014_5484, Hinton2015DistillingTK}, quantization~\cite{DBLP:journals/corr/GongLYB14, Wu_CVPR2016}, or low-rank factorization and sparsity~\cite{Jaderberg_BMVC2014, 6866160}. Another main challenge for DNNs is to remain efficient on data coming from a target distribution similar yet not identical to the training data. A popular technique is to use a model pre-trained on a large source dataset and fine-tune its parameters on the desired target dataset. However it may be that no labels are available on the target data. In that case, training requires unsupervised DA techniques~\cite{Ben-David2010, DANN_Ganin_JMLR2016, MaximumClassifierDiscrepancy_Saito_CVPR2018}.

\begin{SCfigure}[1.22][t]
	\centering
	\includegraphics[width=0.42\linewidth,trim={20mm 0 20mm 0},clip]{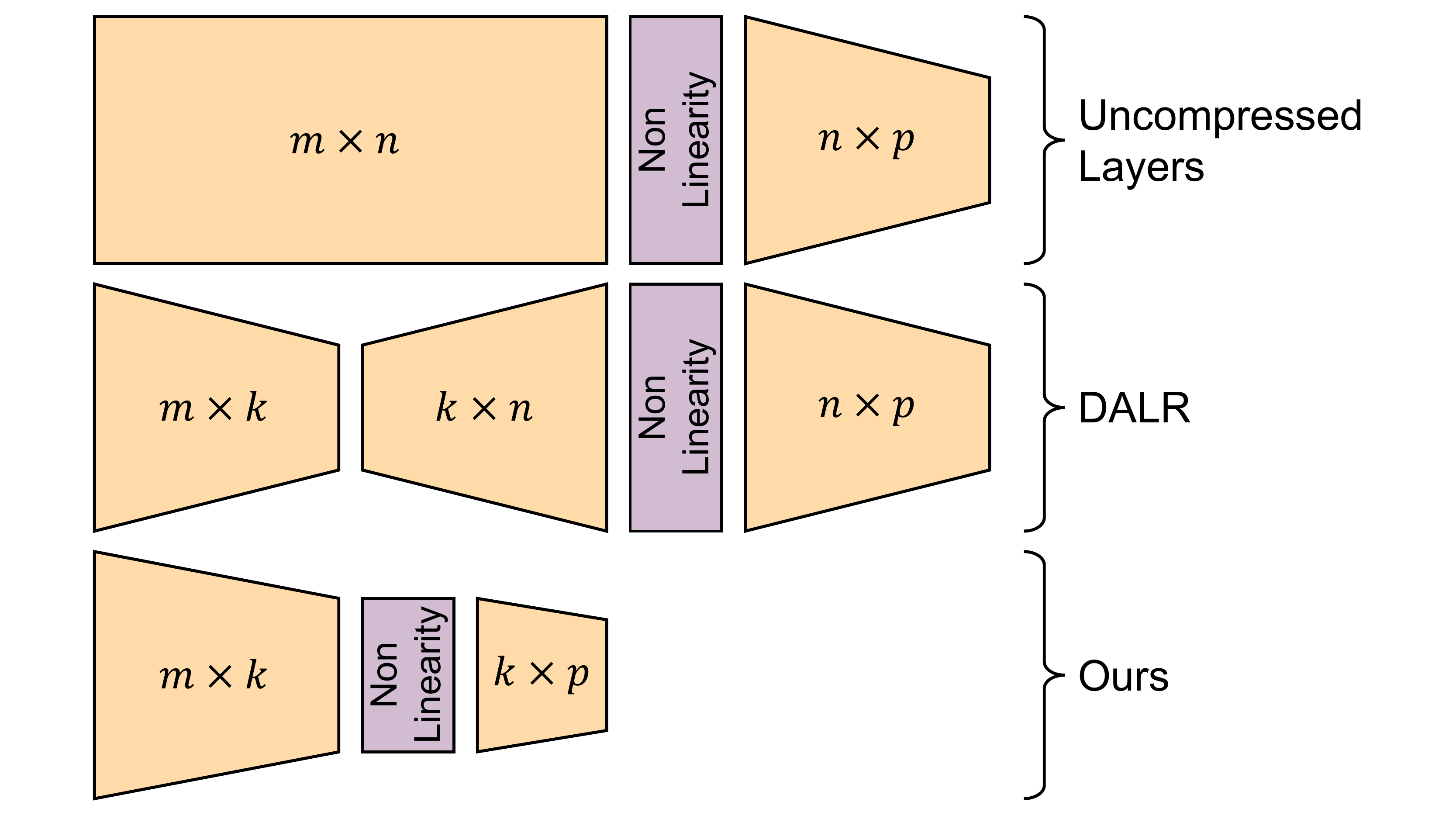}
	\caption{
		We propose a compression method with additional regularization terms for Domain Adaptation (DA) setting.
		Our method can maintain accuracy without fine-tuning after compression,
		and can achieve higher accuracy than prior work (DALR~\cite{DALR_Masana_ICCV2017}).
		DALR is based on low-rank matrix decomposition,
		while our method is built on Spectral Pruning~\cite{SpectralPruning_Suzuki2018}.
	}
	\label{fig:DALR_and_ours}
\end{SCfigure}

In this work, we seek to adapt compression objectives to the specific setting of DA. Concretely, we are provided with both a source and a, possibly unlabeled, target distribution. We are interested in compressing DNNs with high accuracy on the target distribution.
Some works have investigated this specific setting. The work in~\cite{DALR_Masana_ICCV2017} is strongly related to ours but only focuses on the case where a model is fine-tuned on a target dataset. Their method yields great results in this setting because it is data-dependent. That is, compression depends on the input data distribution. 
In this paper, we focus on a compression method that is also data-dependent. The novelty of our work is to go further in the analysis of how the data affects compression and to add a new regularization term. This regularization term is directly related to a DA objective. It basically favors nodes that are non domain discriminative.  This proves useful as the more similar the features extracted are, between source and target distribution, the more discriminative ability learned on the source distribution will also apply to the target distribution. Finally we show that our extended compression method compares favorably to~\cite{DALR_Masana_ICCV2017} on various cases of fine-tuning a pre-trained model on a target dataset.

\section{Related work} \label{section_related_work}

Here we present a brief review of the main strategies used in model compression.
One can find in~\cite{Cheng2018_ModelCompression} a more in-depth analysis of the different methods. 
\textbf{(1) Pruning.}
These methods focus on finding weights of the network that are not crucial to the network's performance. Early works used magnitude-based pruning which removed weights with the lowest values~\cite{NIPS1988_156}. This method was extended by various works yielding significant results~\cite{NIPS2016_6165, ijcai2018-330}.
In this work we focus on a pruning method~\cite{SpectralPruning_Suzuki2018}, which is based on spectral analysis of the covariance matrix of layers~\cite{Suzuki_AISTATS2018, Suzuki_ICLR2020}.
\textbf{(2) Network distillation.}
These approaches are based on a teacher-student learning scheme~\cite{conf/kdd/BucilaCN06, NIPS2014_5484, Hinton2015DistillingTK}. A compressed, small network will learn to mimic a large network or ensemble of large networks. This is efficient because the student model learns based on a combination of true labels, which are hard encoded, and soft outputs of the teacher model. Those contain valuable information about similarity between classes. Distillation can also train the student network to mimic the behavior of intermediate layers~\cite{FitNet}.
\textbf{(3) Parameter quantization.}
Network quantization is based on finding an efficient representation of the network's parameters by reducing the number of bits required to represent each weight. Various strategies can achieve this objective:~\cite{DBLP:journals/corr/GongLYB14, Wu_CVPR2016} applied k-means scalar quantization to the weight values. Other work use a fixed (8 or 16) bit representation of the weights while maintaining the accuracy of the original model~\cite{37631, pmlr-v37-gupta15}.  In~\cite{BinaryConnect} a network is constrained to use only binary weights while still achieving high accuracy.
\textbf{(4) Low-rank factorization.}
The idea behind low-rank factorization and sparse techniques is to decompose layers into smaller ones with fewer parameters. For convolutional layers, main techniques rely on the fact that performing a convolution with the original filters can be approached by a linear combination of a set of convolutions with base filters~\cite{6866160, Denton_NIPS2014}. In~\cite{DALR_Masana_ICCV2017} truncation of Singular Value Decomposition (SVD) is used to factorize the weight matrices of dense (fully connected) layers.

Some works explore pruning methods in DA (or transfer learning) setting~\cite{TaylorPruning_Molchanov_ICLR2017, FinePruning_Tung_BMVC2017, SparseTargetNet_Liu_AAAI2017, PFA_Suau2018, NwA_Zhong_ECCVW2018, TCP_Yu_IJCNN2019}.
Unlike our work, most existing methods depend on iterative fine-tuning.
Although each fine-tuning step may mitigate information loss caused by each pruning step,
strong dependence on fine-tuning increases the risk of overfitting
(see \cite{WealthAdapt} for more discussion).

In contrast to the iterative fine-tuning approach,
the Domain Adaptive Low Rank Matrix Decomposition (DALR) method~\cite{DALR_Masana_ICCV2017}
can maintain accuracy without fine-tuning after compression.
DALR and our method are data-dependent, which makes them suited for DA setting.
Since DALR uses SVD, both are based on spectral analysis.
An important difference between DALR and our method is their structural modification.
We describe the difference, considering that uncompressed two layers have weight matrices of dimension $m \times n$ and $n \times p$ as shown in Fig.~\ref{fig:DALR_and_ours}.
For compressing the first layer, DALR produces two smaller layers that have weight matrices of dimension $m \times k$ and $k \times n$ ($k$ is kept rank).
First, it is noted that, to actually reduce the number of parameters in the network, the parameter $k$ must verify the following condition: $ k(m+n) < mn $. DALR affects only the layer being compressed since the output shape remains unchanged. Conversely, our method does not create additional layers but also affects the input dimension of the following layer.
Second, DALR is only designed to compress dense layers while our method can be applied to both dense and convolutional layers.
Finally,
in~\cite{DALR_Masana_ICCV2017}, the authors only mention using $k$ as a given input to compression.
Therefore compression and validation with various $k$ will be needed
unless a prior knowledge of how much the layer is compressible is known.
Our method helps to avoid this issue, using an information retention ratio parameter defined in Section~\ref{sec:method_principle}.
It should be noticed however that it is possible to reproduce a similar parametrization in DALR by computing the ratio of kept singular values over all singular values of the decomposition.

\section{Spectral pruning with DA regularization} \label{section_method}

Our work was built on a layer-wise compression method~\cite{SpectralPruning_Suzuki2018},
which is based on spectral analysis.
The method does not require any regularization during training, but is applicable to a pre-trained model. 
We will first detail this method then introduce our DA regularizer.

\subsection{Principle}
\label{sec:method_principle}

The method compresses the network layer per layer and can be applied to both dense and convolutional layers. It aims at reducing the dimension of the output of a layer such that the information of the input of the next layer remains mostly unchanged.
To compress a layer, it selects the most informative nodes (neurons) or feature map channels (corresponding to convolutional filters).
See Fig. 3 in~\cite{LWC_Han_NIPS2015} and Fig. 2 in~\cite{ChannelPruning_He_ICCV2017} for visual explanations of pruning dense and convolutional layers, respectively.
For brevity, we will avoid the distinction of the layer types in the remainder of this paper and refer only to nodes for both cases.

More formally, let us denote $x$ the input to the network
and suppose that we are given a training data $(x_i,y_i)_{i=1}^n$ with size $n$. 
Let us consider a model that is a sequence of convolutional and fully connected layers. We denote by $f^l$ and $\eta^l$ respectively the mapping function and activation function of a layer $l$. Finally we denote by  $\phi^l(x) \in \Real^{\mell}$ the representation obtained at layer $l$
where $\mell$ is the number of nodes in the layer $l$. Therefore we have 
$\phi^l(x) = \eta^l(f^l(\phi^{l-1}(x))).$
The objective of the method is to find for each layer $l$, the subset of nodes $J \subset \{1,\dots, \mell\}$ that minimizes the following quantity: 
\begin{align} 
\hat{\E}[\|\phi^l(x) - \hat{A}_J\phi^l_J(x)\|^2],
\label{eq2}
\end{align}
where $\hat{\E}[.]$ denotes the empirical expectation, $\phi^l_J(x)$ denotes the vector composed of the components of $\phi^l(x)$ indexed by $J$, and $\hat{A}_J$ is a matrix to recover the whole vector $\phi^l$ from its sub-vector $\phi^l_J$ as well as possible.

For a fixed $J$, it is easy to minimize the objective \eqref{eq2} with respect to $\hat{A}_J$ as 
$\hat{A}_J = \hat{\Sigma}_{F,J}\hat{\Sigma}_{J,J}^{-1}$,
where $\hat{\Sigma} :=  \hat{\E}[\phi^l(x)\phi^l(x)^\top] (= \frac{1}{n}\sum_{i=1}^n \phi^l(x_i) \phi^l(x_i)^\top)$ is the empirical covariance matrix and $F := \{1,\dots,\mell\}$ is the full set of indexes. 
Then $\phi^l(x)$ can be approximated by $\hat{A}_J\phi^l_J(x)$ which is equivalent to pruning the nodes of layer $l$ that do not belong to $J$ and reshaping the weights of the next layer. Let us denote $W_{l+1}$ the weight matrix if the next layer is a fully connected layer, $T_{l+1}$ the weight tensor if it is a convolutional layer. In that case we denote by the four dimensional tuple $(o_c, i_c, W, H)$ the shape of the weight tensor where $o_c, i_c, W, H$ stand respectively for output channels, input channels, width, height.
The reshaping of the weights of the next layer is done the following way.
For fully connected layers, we obtain the reshaped weight matrix as
$\widetilde{W}^{l+1} = W^{l+1}\hat{A}_J.$
For convolutional layers, it is done by first reshaping $T_{l+1}$ to a $(W,H,o_c,i_c)$ tensor. Therefore each point of the filter grid is associated to a filter matrix $M_{w,h} \in \mathbb{R}^{o_c \times i_c}$ that will be reshaped by the operation:
$\widetilde{M}_{w,h} = M_{w,h}\hat{A}_J.$
The tensor is then reshaped back to its original form $(o_c, i_c, W, H)$ where $i_c$ has been modified by the process.

Once we obtained the optimal $\hat{A}_J$ for a fixed $J$,
then we minimize the objective \eqref{eq2}
with respect to $J$.
By substituting the optimal $\hat{A}_J$ to the objective function \eqref{eq2},
it can be re-written as
$\min_J \Tr[\hat{\Sigma}_{F,F}-\hat{\Sigma}_{F,J}\hat{\Sigma}_{J,J}^{-1}\hat{\Sigma}_{J,F}].$
The minimand in the right hand side is zero for $J =F$ and $\Tr[\hat{\Sigma}_{F,F}]$
for $J = \emptyset$. 
Hence we may consider the following ``ratio'' of residual as an information retention ratio $r$ to measure the relative goodness of $J$:
\begin{equation} \label{eq8}
    r = \frac{\Tr(\hat{\Sigma}_{F,J}\hat{\Sigma}_{J,J}^{-1}\hat{\Sigma}_{J,F})}{\Tr(\hat{\Sigma}_{F,F})}.
\end{equation}
The higher the ratio the more information computed by the layer will be retained. %
It is no greater than 1 since the denominator is the best achievable value with no cardinality constraint. Compression can therefore be parametrized by a required information retention ratio parameter $\alpha$ and the final optimization problem translates to
$\min_{J \subset F} |J| \ \  s.t. \ \ r \geq \alpha.$
To solve the optimization problem, the authors of~\cite{SpectralPruning_Suzuki2018} proposed a greedy algorithm where $J$ is constructed by sequentially adding the node that maximizes the information retention ratio.

It is worth noticing that though the method is aimed at compressing each layer sequentially, using $\alpha$ allows more flexibility than methods with a fixed compression rate for each layer. Since we do not impose constraints on the cardinality of $J$, compression will adapt to the inherent compressibility of the layer. We argue that this makes $\alpha$ an easy to tune and intuitive hyperparameter and, if necessary, can easily be combined with cardinality constraints.

\subsection{Domain Adaptation regularization}

In DA setting, informative nodes could be different between source and target domains.
To adjust this difference, we propose an additional regularization term to select better nodes. 

As shown by~\cite{Ben-David2010}, one of the main challenges of DA is to find a representation discriminative with respect to the classification task but not discriminative between domains. The general idea is that, by matching the two distributions, the classification knowledge learned on the labeled source data can be transferred to the target data. Therefore, we propose to use a measure of the alignment of the source and target features distributions for selecting nodes during compression.
We call our method \textit{\OurAlgoFull (\OurAlgo)}.

Previous work involving covariance alignment achieved high accuracy in DA tasks~\cite{Sun_AAAI2016}. Following those observations, we define our regularization term as the following 
\begin{equation}
\begin{split}
     R_{J} = \|\hat{\E}_{x \sim X^s}[\phi^l_J(x)]-\hat{\E}_{x \sim X^t}[\phi^l_J(x)]\| + \|S_{J,J}\odot(C^s_{J,J} - C^t_{J,J})\|_{\mathrm{F}},
     \label{eq:RJ_indexset}
\end{split}
\end{equation}
where $X^s, X^t$ denotes the source and target distributions, $C^s, C^t$ respectively the source and target empirical covariance matrices, $S$ a scaling matrix defined by $S_{ij} = (C^t_{i,i} C^t_{j,j})^{-\frac{1}{4}}$, $\odot$ denotes the element wise multiplication, and $\|\cdot\|_{\mathrm{F}}$ means the Frobenius norm of a matrix. 
This quantity measures the discrepancy between the source and target distributions
by using the up-to second order statistics.
The first term measures the discrepancy of mean and the second term measures the discrepancy of the (scaled) second order moment.
Hence, this regularization term induces a {\it moment matching} effect and the two distributions become similar on the selected nodes.
We may use the MMD criterion (kernel-based discrepancy measure)~\cite{gretton2012kernel} to capture not only the first and second moments but also all higher moments instead, 
but we found that 
it takes too much computational time.
Although the kernelized alignment loss~\cite{SoHoT_Koniusz_CVPR2017} could speed-up computation, source and target samples per class need to be few.

Instead of the criterion \eqref{eq:RJ_indexset}, 
we also consider the following alternative formulation: 
\begin{equation}
\begin{split}
     R_j = \|\hat{\E}_{x \sim X^s}[\phi^l_j(x)]-\hat{\E}_{x \sim X^t}[\phi^l_j(x)]\| + \|S_{j,F}\odot(C^s_{j,F} - C^t_{j,F})\|.
\end{split}
\end{equation}
The first formulation \eqref{eq:RJ_indexset} depends on the whole subset $J$ while the second one computes information more specific to each node. The second formulation is more computationally efficient since the $R_j$'s can be computed only once for each candidate index. The first one needs to recompute it for every candidate index at each step of the procedure because of its dependence on $J$.

\subsection{Intuition behind our regularization}

The intuition behind our regularization is the following. In DA setting where few or no labels are available on the target distribution, the discriminative knowledge is learned mainly on the source distribution.
Thus the feature distributions on both domains should be aligned for the discriminative knowledge learned on the source domain to apply to the target domain.
When comparing the information retention ratios of Eq. \ref{eq8} for different nodes, we observed that many of them had very close values. This means that many nodes capture approximately the same amount of information, in the sense of total variance. Our intuition was that when these relative differences are too small, they are not significant to discriminate between nodes and are more likely the result of noise. Therefore a better criterion should be used to differentiate nodes that capture a same amount of variance. Since our compression method is designed to be applied on models that have been trained using DA methods, it is natural to use a criterion related to DA objectives. We choose to compare nodes based on how well their feature distributions on source and target domains are aligned. Nodes leading to a better alignment should be favored as they will allow for a better transfer of discriminative knowledge from source to target domain.
Our method realizes this via a moment matching type regularization.

\subsection{Practical implementation}

\vspace{-3mm}
\setlength{\columnsep}{-38pt}
\begin{multicols}{2}
\noindent
We present in Algorithm~\ref{alg1} a pseudo code for a practical implementation of our method using the first formulation of our regularizer. If the second formulation is used, the regularization term for each node is computed only once before entering the while loop as it does not depend on $J$. At each step of the algorithm, we denote by $C$ the set of candidate indexes. To select which node to add to $J$, the ratio from \Eqref{eq8} is computed for each candidate index.  We denote by $V$ the vector where each coordinate corresponds to the ratio value of a candidate index. Similarly, we denote by $R_J$ the vector where each coordinate correspond to the regularizing term associated to a candidate index.

Without regularization, the index to add to $J$ is
\columnbreak

\hspace{10mm}
\scalebox{0.85}{
\begin{minipage}{0.89\linewidth}
\vspace{-1mm}
\setlength{\algomargin}{1mm}
\begin{algorithm}[H]
	\setstretch{0.7}
	\SetAlCapHSkip{0mm}
	\DontPrintSemicolon
	\footnotesize
	\caption{\OurAlgo}
	\SetKwInOut{Input}{Input}
	\SetKwInOut{Output}{Output}
	\SetAlgoLined
	\SetKwProg{Fn}{Function}{}{end}
	\Input{$\hat{\Sigma}_{F,F}$ : Empirical covariance matrix,\\ $\alpha$ : Required information retention ratio
	}
	\Output{$J$ : Subset of selected nodes
	}
	\Fn{FindSubset$(\hat{\Sigma}_{F,F}, \alpha)$}
	{
		
		$J \leftarrow \emptyset $\\
		$T \leftarrow \Tr(\hat{\Sigma}_{F,F})$\\
		$r \leftarrow 0$\\
		$ C \leftarrow F$ \\
		\While{$r < \alpha$}{
			\For{$j \in C$}{
				$J_j \leftarrow J\cup j$\\
				Compute $R_{J_j}$\\
				$V_j \leftarrow \frac{\Tr(\hat{\Sigma}_{F,J_j}\hat{\Sigma}_{J_j,J_j}^{-1}\hat{\Sigma}_{J_j,F})}{T}$
			}
			$i \leftarrow \argmaxunder{j \in C} \left \{ V_j -  \lambda \cdot \mathrm{\sigma}(V) \cdot \frac{R_{J_j}}{\max_{j'}(R_{J_{j'}})}\right \}$\\
			$J \leftarrow J\cup \{i\}$\\
			$r \leftarrow V_i$\\
			$C \leftarrow C\setminus\{i\}$
		}
	}
	
	\label{alg1}
\end{algorithm}
\end{minipage}%
}
\end{multicols}
\vspace{-4mm}

\noindent
simply given by
$i = \argmax_{j \in C} V_j.$
Using regularization, the index to add to $J$ is chosen as
\begin{equation}
i = \argmax_{j \in C} \left \{ V_j - \lambda \cdot \mathrm{\sigma}(V)\cdot \frac{R_{J_j}}{\max_{j'}(R_{J_{j'}})}\right \},
\end{equation}
where $\sigma(V)$ is the standard deviation of $(V_j)_{j \in C}$.
The values of $R_{J_j}$ are rescaled to be in the same range as $V_j$'s by max normalization. Multiplying by $V$'s standard deviation, $\sigma(V)$, ensures that the regularization will be applied at the right scale. It should only slightly modify the relative ordering of the values in $V$ as the main criteria for selection must remain the information ratio. Indeed, only taking into account the regularization term would favor nodes that output a constant value across both distributions.
The hyperparameter $\lambda$ allows for more control over the trade-off between those two terms.
We use $1$ as the default $\lambda$ in our experiments.
The max normalization and scaling make $\lambda$ easier to tune. This compression is applied after training and can therefore be combined with any DA method.
\vspace{-1mm}

\section{Experiments on digits images} \label{section_experiments_digits}
In this section, we conduct experiments with a model trained on digits datasets, using a DA technique~\cite{MaximumClassifierDiscrepancy_Saito_CVPR2018}. The source dataset is the SVHN dataset~\cite{37648} and the target dataset is the MNIST dataset~\cite{Lecun1998},
which are standard for DA with adversarial training~\cite{DANN_Ganin_JMLR2016,MaximumClassifierDiscrepancy_Saito_CVPR2018,ADDA_Tzeng_2017_CVPR}.

\subsection{Model}
Considering the relative simplicity of the task, we used a custom model composed of a 4 layers feature generator (3 convolutional + 1 dense) and a 2 layers classifier (see \AppendixSection for details).
To train the model we used the DA technique presented in~\cite{MaximumClassifierDiscrepancy_Saito_CVPR2018} which utilizes ``maximum classifier discrepancy." Briefly, the adaptation is realized by adversarial training of a feature generator network and two classifier networks. The adversarial training forces the feature generator to align the source and target distribution. Contrary to other methods relying on adversarial training with a domain discriminator, this method considers the decision boundary while aligning the distributions, ensuring that the generated features are discriminative with respect to the classification task at hand.

\noindent
\begin{minipage}[t]{.48\textwidth}
	\vspace{-4.0mm}
	\begin{figure}[H]
		\centering
		\includegraphics[width=0.75\linewidth]{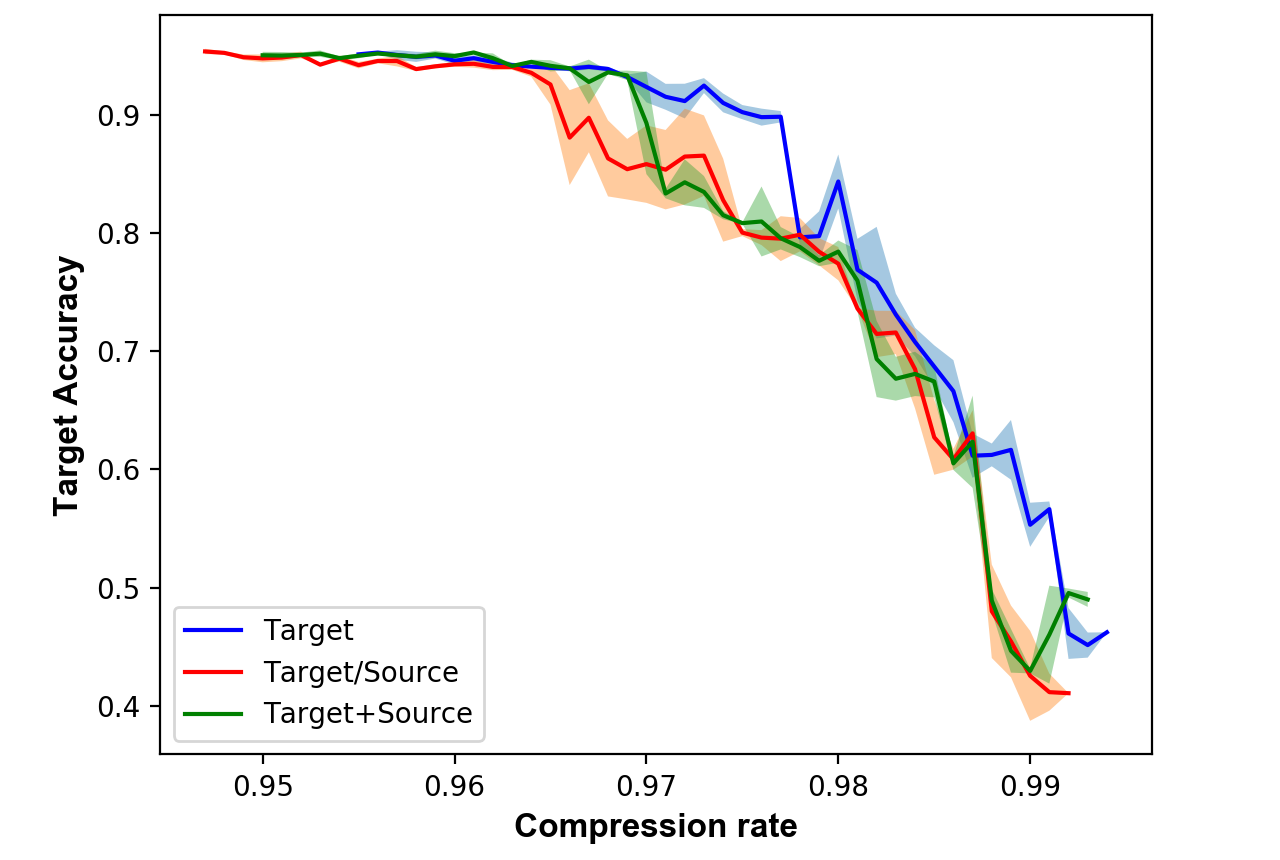}
		\vspace{-4mm}
		\caption{Target accuracy over compression rate for three different settings of data used as input. Results are presented as mean +/- standard deviation based on 10 iterations.}
		\label{fig:digits1}
	\end{figure}
	\vspace{-5mm}
	\begin{table}[H]
		\setlength{\tabcolsep}{3mm}
		\renewcommand\arraystretch{0.9}
		\setlength\aboverulesep{0.1ex}
		\setlength\belowrulesep{0.4ex}
		\begin{center}
			\small
			\scalebox{0.75}{
				\begin{tabular}{lccc}
					\toprule
					\multirow{2}{*}{Test data} & \multicolumn{3}{c}{First layer nodes specificity} \\
					\cmidrule{2-4}
					& \ Source \   & \ Target \   & \ None \   \\
					\midrule
					Source & 0.79 & 0.77 & 0.72 \\
					\midrule
					Target & 0.65 & 0.84 & 0.71 \\
					\bottomrule
				\end{tabular}
			}
		\end{center}
		\vspace{-2.5mm}
		\caption{Mean activation rates.}
		\label{tab:mean_activation_rates}
	\end{table}
\end{minipage}
\hfill
\begin{minipage}[t]{.48\textwidth}
	\vspace{-4.0mm}
	\begin{figure}[H]
		\centering
		\includegraphics[width=0.75\linewidth]{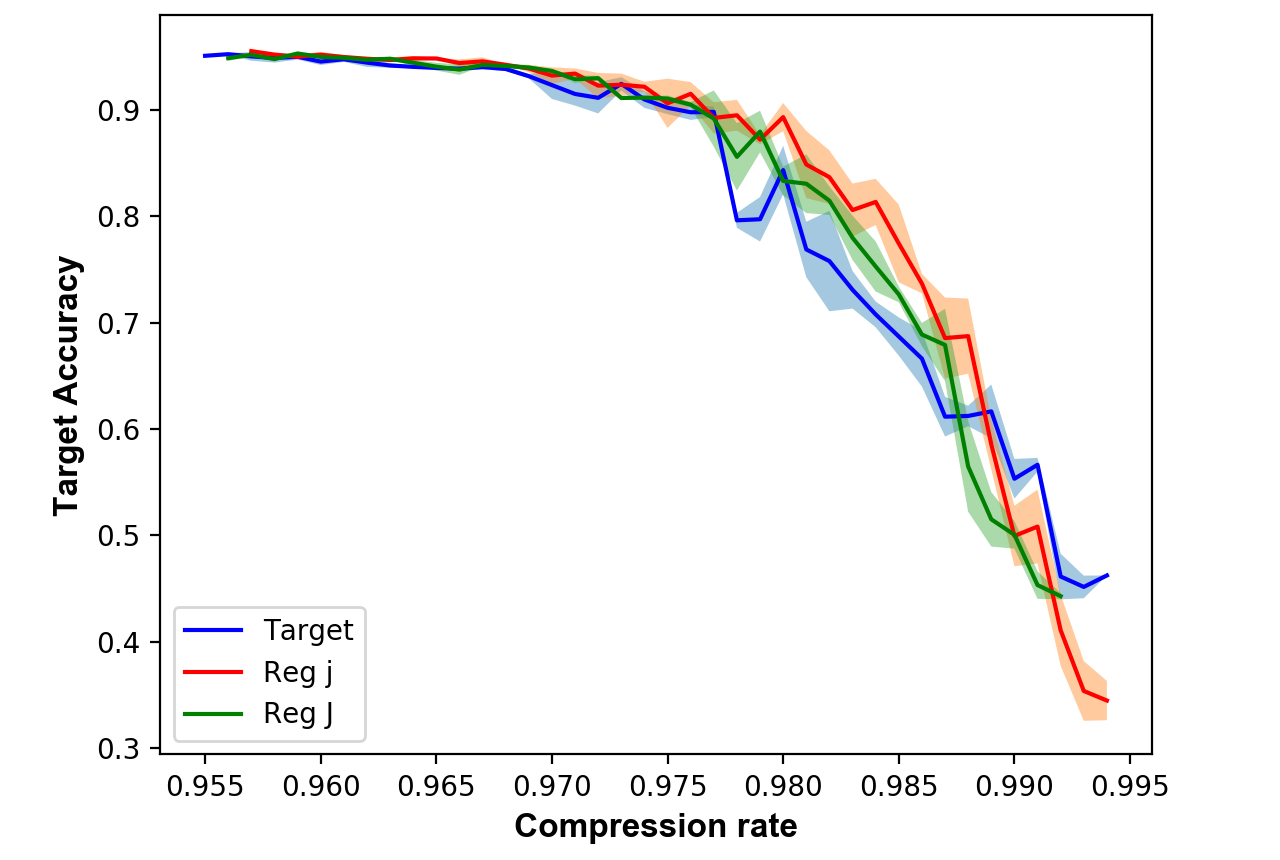}
		\vspace{-4mm}
		\caption{Comparison using only target data or each formulation of our DA regularization. Results are presented as mean +/- standard deviation based on 10 iterations.}
		\label{fig:digits2}
	\end{figure}
	\vspace{-5mm}
	\begin{table}[H]
		\setlength{\tabcolsep}{2.4mm}
		\renewcommand\arraystretch{1.0}
		\setlength\aboverulesep{0.1ex}
		\setlength\belowrulesep{0.4ex}
		\begin{center}
			\small
			\scalebox{0.75}{
				\begin{tabular}{lcccccc}
					\toprule
					C.R. (\%) & 96.0 & 96.5 & 97.0 & 97.5 & 98.0 & 98.5 \\
					\midrule
					Target (\%) & 94.6 & 94.0 & 92.4 & 90.2 & 84.4 & 68.7 \\
					\midrule
					Reg $J$ (\%) & 95.0 & 94.1 & 93.7 & \textbf{91.1} & 83.3 & 72.7  \\
					Reg $j$ (\%) & \textbf{95.2} & \textbf{94.9} & \textbf{94.3} & 90.6 & \textbf{89.3} & \textbf{77.5}  \\
					\bottomrule
				\end{tabular}
			}
		\end{center}
		\vspace{-2.5mm}
		\caption{Accuracy on target test dataset for different compression rate (C.R.).}
		\label{tab:digits_summary_of_compression}
	\end{table}
\end{minipage}
\vspace{-1.0mm}

\subsection{Data choice for compression}
The uncompressed model was trained using both train splits and evaluated on the test split of MNIST. It reached an accuracy of 96.64\%, similar to the results obtained in the original paper~\cite{MaximumClassifierDiscrepancy_Saito_CVPR2018}. We then apply compression on the trained model. During compression, only data of the train splits was used and the compressed models were evaluated on the MNIST test split. Since the method is data-dependent, the choice of data to use to compute the empirical covariance matrix should be taken care of. Three different settings were tested:
(1) Target: Using 14,000 samples from the target distribution;
(2) Target/Source: Using 7,000 samples from the target distribution and 7,000 samples from the source distribution; and
(3) Target+Source: Using 14,000 samples from the target distribution and 7,000 additional samples from the source distribution.
The results obtained are presented in Fig.~\ref{fig:digits1}. 
Using only target samples to compute the covariance matrix shows the best performance. 

To give a better understanding of this result, 
we conducted an analysis about how the nodes activation pattern depends on data distribution as follows: 
we compressed the first layer of a trained network using either only target data or only source data. We then compared the activation of nodes that were selected only in one of the two cases, in other words, nodes specific to target or source data.
We show the results in Table~\ref{tab:mean_activation_rates}.

As expected, nodes selected using source data had a significantly higher activation rate on the source distribution
and conversely for target specific nodes.
As a control case, we added activation of nodes that were selected in neither of the two settings.
Those do not show any significant difference in their activation. 
Interestingly, this difference was no longer appearing when comparing activation of the last fully connected layer,
because DA training aligns the distributions of extracted features.
This experiment sheds light on how the input data affects the compression of data-dependent methods. In case of a model trained on different distributions with DA, early layers contain nodes that are specific to each distribution.
Thus it is critical for compression to use data coming exclusively from the distribution the model will be applied to. Partially using source data forces compression in the early layers to select source specific nodes to the expense of target specific nodes leading to poorer results.

\subsection{Regularization for compression}

Finally we compared the results of the best baseline, using only target data, with adding the regularization term introduced in Section~\ref{section_method}. The results are presented in Fig.~\ref{fig:digits2} and summarized in Table~\ref{tab:digits_summary_of_compression}. It appears that the second formulation of our regularizer gives the best performance. This is probably due to the fact it focuses on node level information and not on the whole subset $J$, giving better ability to discriminate against specific nodes.

Compared to the baseline, our regularizer leads to significant improvements in compression performance. We observe up to 9\% bumps in accuracy.
Yet, we notice that the baseline performs slightly better for very high compression rates.
We conjecture that reducing information loss becomes more important for such compression rates
and our regularization with $\lambda{=}1$ is too strong.
There is room for improvement by tuning the hyperparameter $\lambda$ depending on the compression rate or developing methods for adaptive regularization.

\section{Experiments on natural images} \label{section_experiments_natural}
In this section, we compare our method (\OurAlgo) with the factorization based compression method (DALR) in~\cite{DALR_Masana_ICCV2017}. We reproduce the same setting as in their experiment: a VGG19 model pre-trained on the ImageNet dataset~\cite{ImageNet_IJCV2015} is fine-tuned on different target datasets.

\subsection{Experimental settings}

We establish our comparison based on the following three datasets used in~\cite{DALR_Masana_ICCV2017} experiments: Oxford 102 Flowers~\cite{Nilsback2008}, CUB-200-2011 Birds~\cite{WahCUB_200_2011}, and Stanford 40 Actions~\cite{Yao2011HumanAR}.

For all three datasets, we first trained uncompressed models by fine-tuning from an ImageNet pre-trained VGG19 model.
In the fine-tuning before compression, we trained the weights of the fully connected layers while keeping the weights of the convolutional layers frozen to their pre-trained value on ImageNet like~\cite{DALR_Masana_ICCV2017}.

We then compressed the models.
The input data to compression was always composed of 4,000 samples of the train split, except for the Oxford 102 Flowers where the whole train split was used.
Additional 4,000 randomly sampled images from the ImageNet train split were used for DA regularization. 
Note that our method and DALR~\cite{DALR_Masana_ICCV2017} do not need the labels of target datasets in this compression phase.
After that we optionally fine-tuned the fully connected layers of compressed models with target labels.

We used the Adam optimizer~\cite{Adam} with
a batch size of 50,
a learning rate of $10^{-4}$, %
and a weight decay of $5{\times}10^{-4}$ %
for training models.
See \AppendixSection for details.

\subsection{VGG19 fc7 compression}

We first evaluated the compression on the last fully connected layer (fc7) of the model, containing 4,096 nodes,
because the fc7 compression is a main evaluation setting in~\cite{DALR_Masana_ICCV2017}.
For each trial we report the results both with and without fine-tuning of the classifier layers after compression. We also report the results of compression using basic SVD on fc7 weight matrix as a baseline to further illustrate the advantage of using data-dependent methods.
To compare our method with DALR, a dimension $k$ was set for the compression using DALR then $k'$, the dimension to keep in our method resulting in an equal number of parameters, was determined accordingly.
See \AppendixSection for details.

The results are presented in Tables~\ref{tab:VGG19_OxfordFlowers}, \ref{tab:VGG19_CUB200}, and~\ref{tab:VGG19_StanfordActions}.
We show the relative numbers of parameters in the fc7 layer compressed by DALR
in the ``params'' rows in the tables as in the DARL paper~\cite{DALR_Masana_ICCV2017}.
In all experiments, our method maintains high accuracy even for high compression rates. In such cases it outperforms DALR by a large margin. However, for lower compression rates DALR consistently compares favorably to our method though the difference is arguably small.
In most cases, fine-tuning does not improve much the performance of any of the two methods, except for DALR for high compression rates.
This phenomenon implies that the learning rate and/or the dropout rate we used are too high especially for our method (see additional fine-tuning results in \AppendixSection for details).

\begin{table}[t]
\setlength{\tabcolsep}{2.2mm}
\renewcommand\arraystretch{0.9}
\setlength\aboverulesep{0.0ex}
\setlength\belowrulesep{0.5ex}

\begin{minipage}[t]{.49\textwidth}
\begin{center}
\small
\scalebox{0.75}{
\begin{tabular}{lcccccc}
\toprule
 $k$ & $4$ & $8$ & $16$ & $32$ & $64$ & $128$ \\
params (\%) & 0.20 & 0.39 & 0.78 & 1.56 & 3.13 & 6.25 \\ %
\midrule
SVD w/o FT. & \ \ 5.2 & 12.8 & 30.2 & 54.6 & 65.2 & 67.6 \\
SVD w/ FT. & 13.7 & 36.7 & 56.6 & 65.9 & 70.0 & 70.8 \\
\midrule
DALR w/o FT. & \ \ 8.6 & 31.1 & 55.1 & 66.9 & 71.0 & \textbf{72.8} \\
DALR w/ FT. & 18.9 & 48.2 & 64.4 & \textbf{70.3} & 70.7 & \textbf{72.3} \\
\midrule
Ours w/o FT. & \textbf{68.9} & \textbf{69.3} & \textbf{70.0} & \textbf{71.2} & \textbf{72.0} & 72.5 \\
Ours w/ FT. & \textbf{67.3} & \textbf{68.9} & \textbf{70.3} & 69.9 & \textbf{70.8} & 70.6 \\
\bottomrule
\end{tabular}
}
\end{center}
\vspace{-2mm}
\caption{VGG19 fc7 compression on Oxford 102 Flowers. Test dataset accuracy results (\%). Original accuracy: 73.1\%.}
\label{tab:VGG19_OxfordFlowers}
\end{minipage}
\hfill
\begin{minipage}[t]{.49\textwidth}
\begin{center}
\small
\scalebox{0.75}{
\begin{tabular}{lcccccc}
\toprule
 $k$ & $4$ & $8$ & $16$ & $32$ & $64$ & $128$  \\
params (\%) & 0.20 & 0.39 & 0.78 & 1.56 & 3.13 & 6.25 \\ %
\midrule
SVD w/o FT. & \ \ 4.1 & 12.6 & 30.3 & 46.9 & 54.8 & 60.5 \\
SVD w/ FT. & 17.3 & 39.1 & 51.2 & 57.2 & 58.8 & 58.5 \\
\midrule
DALR w/o FT. & \ \ 5.6 & 23.1 & 49.3 & 57.8 & \textbf{60.1} & \textbf{60.9} \\
DALR w/ FT. & 19.0 & 40.9 & 54.0 & \textbf{59.8} & \textbf{59.3} & 59.5 \\
\midrule
Ours w/o FT. & \textbf{58.3} & \textbf{58.2} & \textbf{58.2} & \textbf{58.8} & 59.3 & 60.2 \\
Ours w/ FT. & \textbf{57.9} & \textbf{57.3} & \textbf{57.7} & 58.2 & 58.2 & \textbf{59.7} \\
\bottomrule
\end{tabular}
}
\end{center}
\vspace{-2mm}
\caption{VGG19 fc7 compression on CUB-200-2011 Birds. Test dataset accuracy results (\%). Original accuracy: 61.5\%.}
\label{tab:VGG19_CUB200}
\end{minipage}
\end{table}

\begin{table}[t]
\setlength\aboverulesep{0.0ex}
\setlength\belowrulesep{0.5ex}
\renewcommand\arraystretch{0.9}
\setlength{\tabcolsep}{2.2mm}
\begin{minipage}[t]{.49\textwidth}
\begin{center}
\small
\scalebox{0.75}{
\begin{tabular}{lcccccc}
\toprule
 $k$ & $4$ & $8$ & $16$ & $32$ & $64$ & $128$  \\
params (\%) & 0.20 & 0.39 & 0.78 & 1.56 & 3.13 & 6.25 \\ %
\midrule
SVD w/o FT. & 17.1 & 31.0 & 54.5 & 67.3 & 72.2 & 72.7 \\
SVD w/ FT. & 45.4 & 62.0 & 68.6 & 72.8 & 73.1 & 73.7 \\
\midrule
DALR w/o FT. & 20.7 & 43.3 & 65.0 & 73.0 & \textbf{74.3} & \textbf{74.7} \\
DALR w/ FT. & 46.8 & 63.3 & 69.6 & \textbf{73.5} & 72.9 & 72.9 \\
\midrule
Ours w/o FT. & \textbf{69.0} & \textbf{70.2} & \textbf{71.8} & \textbf{73.1} & 73.9 & 73.8 \\
Ours w/ FT. & \textbf{70.5} & \textbf{70.4} & \textbf{71.7} & 72.9 & \textbf{73.7} & \textbf{74.0} \\
\bottomrule
\end{tabular}
}
\end{center}
\vspace{-2mm}
\caption{VGG19 fc7 compression on Stanford 40 Actions. Test dataset accuracy results (\%). Original accuracy: 74.6\%.}
\vspace{3mm}
\label{tab:VGG19_StanfordActions}
\end{minipage}
\hfill
\setlength{\tabcolsep}{1.1mm}
\begin{minipage}[t]{.49\textwidth}
	\begin{center}
		\small
		\scalebox{0.75}{
		\begin{tabular}{llllllccc}
			\toprule
			\multirow{2}{*}{} & \multicolumn{2}{l}{Layers} & \multirow{2}{*}{FT.} & \multirow{2}{*}{\shortstack{\\C.R.\\(\%)}} & \multirow{2}{*}{\#params} & \multicolumn{3}{c}{Accuracy (\%)} \\
			\cmidrule{2-3} \cmidrule{7-9}
			& conv & fc & &  &  & Oxford & CUB & Stanford \\
			\midrule
			\multirow{2}{*}{DALR} & & \cmark && \multirow{2}{*}{84.9} & \multirow{2}{*}{21.1 M} & 48.1 & 43.0 & 57.7 \\
			& & \cmark & \cmark &&& 62.2 & 52.8 & 67.4 \\  %
			\midrule
			\multirow{2}{*}{Ours} & & \cmark && \multirow{2}{*}{85.0} & \multirow{2}{*}{21.1 M} & 48.1 & 37.1 & 61.0 \\
			& & \cmark & \cmark &&& 63.9 & 55.1 & 69.0 \\  %
			\midrule
			\multirow{2}{*}{Ours} & \cmark & \cmark && \multirow{2}{*}{\textbf{85.4}} & \multirow{2}{*}{\textbf{20.4 M}} & \textbf{60.6} & \textbf{51.4} & \textbf{69.2} \\  %
			& \cmark & \cmark & \cmark &&& \textbf{71.4} &\textbf{58.3} & \textbf{72.6} \\  %
			\bottomrule 
		\end{tabular}
		}
	\end{center}
	\vspace{-2mm}
	\caption{
		VGG19 full compression on each dataset.
	}
	\label{tab:VGG19_full_compression}
\end{minipage}
\end{table}

\subsection{VGG19 full compression}

It is important to notice that, contrary to DALR, our method is able to compress both convolutional and dense layers. To further demonstrate this advantage, we proceeded to the comparison of fully compressing the VGG19 network using both methods.
We conducted the experiment on all three datasets.
DALR was applied to the fully connected layers.
The dropout layers of models compressed by our method were disabled in fine-tuning,
because our method reduce the input dimension of the dropout layers unlike DALR, resulting in performance degradation.
See \AppendixSection for other details.

Results are presented in Table~\ref{tab:VGG19_full_compression}.
For all three datasets, our method consistently achieves a better compression rate (C.R.) while reaching test accuracy 5.2\%--12.5\% higher than DALR.
If we apply our method to the fully connected layers only, the difference in accuracy is smaller but still favorable to our method in many cases.
Although further improvement would be preferable to keep representations for many classes (CUB-200 without fine-tuning), in practice, it is not necessary to overcompress the fully connected layers by our method.

\subsection{Discussion}

Our method is drastically better for high compression rates, while DALR is slightly better for low compression rates.
We conjecture that this is mainly due to the fact that the two methods modify differently the network (see Fig.~\ref{fig:DALR_and_ours}).
DALR affects only one layer. Therefore if the compression is too important, it reaches a critical point where not enough information can possibly be retained.
Our method affects two successive layers therefore spreading the effect of compression and avoiding this critical point for high compression rates.
On the other hand, our method needs to consider nonlinearity between two layers and uses a greedy algorithm~\cite{SpectralPruning_Suzuki2018}.
Thus only affecting one layer is better to maintain high accuracy for lower compression rates,
because the output of the one layer can be optimally approximated.

The optimization problem to solve in DALR admits a closed form solution, making it a very fast compression method.
Compared to DALR, it is a limitation of our method to require an iterative optimization.
However, the extra computational time is usually a few hours on 1 GPU (Tesla V100) in our experiments.
Furthermore, an iterative process is used in the DALR paper~\cite{DALR_Masana_ICCV2017} to determine compression rate pairs for fc6 and fc7,
and it needs iterative accuracy evaluation.
Pruning methods based on iterative fine-tuning approach also take time for pruning, fine-tuning, and evaluation.
Therefore our method is practical enough.

\section{Conclusions} \label{section_conclusions}

In this paper, we investigated compression of DNNs in the DA setting. As shown by~\cite{DALR_Masana_ICCV2017}, using a data-dependent method is crucial in order to achieve good results.
In that matter, our work shows that the input data to calculate information retention ratio for compression should only come from the distribution on which the model will be applied to, the target distribution in our case.
This is because adding samples from another distribution will force compression in the early layers to select nodes that are specific to this distribution.
However, we show that source data can still be used for regularization to improve nodes selection. This is done by comparing the first and second order statistics of the node's feature distributions on each of the source and target data.
This criterion serves as a measure of the alignment of the two distributions which directly relates to DA objectives. Therefore we denote this measure as a DA regularizer.
We evaluated this regularization on a spectral pruning method introduced in~\cite{SpectralPruning_Suzuki2018} and obtained significant improvements on its compression results.
Finally we compared our regularized compression method with the factorization based method of~\cite{DALR_Masana_ICCV2017} on real world image datasets.
Our method compares favorably on all three datasets, leading to significant improvements in retained accuracy for high compression rates.

Although our work focused on one compression method,
we argue that using first and second order statistics of feature distributions to measure the alignment between source and target features and using it as a criterion for compression can be applied to other methods.
This work can therefore serve as a first example and practical implementation of this idea.
Applying our method to unsupervised DA setting~\cite{TCP_Yu_IJCNN2019, OpenMIC_Koniusz_ECCV2018} is an interesting future direction.

\paragraph*{Acknowledgements}
TS was partially supported by JSPS KAKENHI (26280009, 15H05707, and 18H03201), Japan Digital Design, and JST-CREST.

\bibliography{dasp}

\clearpage
\appendix
\section*{\AppendixSection}

\section{Experimental settings on digits images}

The output widths of the layers of the custom model are $(64, 64, 128, 1024, 1024, 10)$.
Each convolutional or dense layer except for the output layer is followed by a batch normalization layer and a ReLU layer, and the feature generator is followed by a dropout layer.

\section{Experimental settings on natural images}

We establish our comparison based on three datasets used in~\cite{DALR_Masana_ICCV2017} experiments.

\vspace{-0.3\baselineskip}
{
	\setlength{\leftmargini}{13pt} 
	\begin{itemize}
		\setlength{\itemsep}{-5.0mm}
		\setlength{\labelsep}{3pt}
		\item \textbf{Oxford 102 Flowers}~\cite{Nilsback2008}: contains 8,189 images. Train and validation splits contain each 1,020 samples with each class equally represented. Test split contains 6,149 samples with classes not equally represented.
		\\
		\item \textbf{CUB-200-2011 Birds}~\cite{WahCUB_200_2011}: contains 11,788 images (5,994 train, 5,794 test) of 200 bird species.
		Although each image is annotated with bounding box, part location (head and body), and attribute labels, those annotations were not used in our experiments.
		\\
		\item \textbf{Stanford 40 Actions}~\cite{Yao2011HumanAR}: contains 9,532 images (4,000 train, 5,532 test) of 40 categories corresponding to different human actions. Classes are equally represented on the training set and all samples contain at least one human performing the corresponding action.
	\end{itemize}
}
\vspace{-0.3\baselineskip}

We used an ImageNet pre-trained VGG19 model that provided by the torchvision package of PyTorch.
We trained for
10 epochs on Oxford-102 Flowers,
5 epochs on CUB-200 Birds,
and 5 epochs on Stanford 40 Actions
for fine-tuning before compression.
On each dataset,
we trained five models to mitigate randomness,
and used the model that has the median accuracy of the five models as the original (uncompressed) model.
When we fine-tune models after compression, we trained for 2 epochs and 5 epochs for VGG19 fc7 compression and VGG19 full compression, respectively.
The dropout rate of VGG19 was set to 0.5.
All models were trained using PyTorch.

In the case of VGG19 fc7 compression,
compression rates for the total parameters of VGG19 models are limited to 11\%--12\% and FLOPs reduction is negligible,
because layers before the fc7 layer are not compressed.

To compare our method with DALR, we need a way to fit the numbers of parameters
because the two methods do not modify the architecture of the network the same way. DALR replaces a fully connected layer by two smaller layers without changing the next layer. Our method keeps the same number of layers but also affects the input dimension of the next layer. Therefore we proceeded the following way to compare the two methods objectively. A dimension $k$ was set for the compression using DALR then $k'$, the dimension to keep in our method resulting in an equal number of parameters, was determined accordingly.
More precisely, the fc7 layer has a weight matrix of dimension $m\times m$
(because $m$ and $n$ in Fig.~1 are the same value for the fc7 layer)
and the next layer has a weight matrix of dimension $m\times p$.
Taking into account the biases we get the following equation: 
\setcounter{equation}{5}
\begin{align}
k' &= \frac{m(2k+1+p) + k}{m+1+p}.
\end{align}

The iterative process of the DALR paper~\cite{DALR_Masana_ICCV2017} for determining compression rate for each layer is computationally inefficient.
Thus, for VGG19 full compression, we did not determine compression rate for each layer automatically for both methods.
Specifically, for compression by DALR,
compression rate for fc8 is set to 0.5 (a modest value for not breaking output values), and
compression rate for other layers (fc6 and fc7) is set so that total compression rate becomes $\sim$85\%.
For our method,
compression rate for fully connected layers is set to 0.96, and
compression rate for convolutional layers is set so that total compression rate becomes over 85\%.
Although the FLOPs reduction by DALR is negligibly small,
our method reduces FLOPs by $\sim$19\% thanks to the compression of convolutional layers.

\section{Fine-tuning results on natural images}

In many cases,
fine-tuning degrades the performance of our method.
To investigate this phenomenon, we evaluated other models that fine-tuned for various epochs on Oxford 102 Flowers.
The results are presented in Table~\ref{tab:VGG19_OxfordFlowers_FT5epoch}.
The accuracy drops in the first epoch and recovers in succeeding epochs.
Considering these results, the learning rate and/or the dropout rate we used are too high for our method,
and they cause the performance drops by keeping parameters away from optimal values.
Tuning learning rates, dropout rates, and epochs for fine-tuning by cross-validation will further improve accuracy, though it takes much time.

\setcounter{table}{6}
\begin{table}[t]
	\setlength{\tabcolsep}{2mm}
	\renewcommand\arraystretch{0.85}
	\setlength\aboverulesep{0.0ex}
	\setlength\belowrulesep{0.5ex}
	\begin{center}
		\small
		\begin{tabular}{lcccccc}
			\toprule
			$k$ & $4$ & $8$ & $16$ & $32$ & $64$ & $128$ \\
			\midrule
			Ours w/o FT.      & 68.9 & 69.3 & 70.0 & 71.2 & 72.0 & 72.5 \\
			Ours 1 epoch FT.  & 64.0 & 63.9 & 64.9 & 67.7 & 69.5 & 67.8 \\
			Ours 2 epochs FT. & 67.3 & 68.9 & 70.3 & 69.9 & 70.8 & 70.6 \\
			Ours 5 epochs FT. & \textbf{71.9} & \textbf{72.9} & \textbf{73.2} & \textbf{73.5} & \textbf{74.2} & \textbf{74.2} \\
			\bottomrule
		\end{tabular}
	\end{center}
	\vspace{-2mm}
	\caption{
		Accuracy transition by fine-tuning. VGG19 fc7 compression on Oxford 102 Flowers.
		Test dataset accuracy results (\%).
	}
	\label{tab:VGG19_OxfordFlowers_FT5epoch}
\end{table}

\end{document}